\documentclass[conference]{IEEEtran}

\usepackage[dvips]{graphicx}
\usepackage{float}
\usepackage{times}
\usepackage{amsmath}
\usepackage{amsfonts}
\usepackage{balance}
\usepackage{multirow}

\begin{document}
%
\title{A Bayesian Compressed Sensing Kalman Filter for Direction of Arrival Estimation}

\author{\IEEEauthorblockN{Matthew Hawes$^{a}$, Lyudmila Mihaylova$^{a}$, Francois Septier$^{b}$  and Simon Godsill$^{c}$}\\[0.3cm]
\IEEEauthorblockA{$^{a}$ Department of
Automatic Control and Systems Engineering, University of Sheffield, S1 3JD, UK\\
$^{b}$ Institute Mines Telecom/Telecom Lille, CRIStAL UMR CNRS 9189, France  \\
$^{c}$ Department of
Engineering, Cambridge University, CB 1PZ, UK\\
{\{m.hawes, l.s.mihaylova\}@sheffield.ac.uk},
francois.septier@telecom-lille.fr, sjg30@cam.ac.uk}

}
\maketitle

\begin{abstract}
In this paper, we look to address the problem of estimating the dynamic direction of arrival (DOA) of a narrowband signal impinging on a sensor array from the far field.  The initial estimate is made using a Bayesian compressive sensing (BCS) framework and then tracked using a Bayesian compressed sensing Kalman filter (BCSKF).  The BCS framework splits the angular region into $N$ potential DOAs and enforces a belief that only a few of the DOAs will have a non-zero valued signal present.  A BCSKF can then be used to track the change in the DOA using the same framework.  There can be an issue when the DOA approaches the endfire of the array.  In this angular region current methods can struggle to accurately estimate and track changes in the DOAs.  To tackle this problem, we propose changing the traditional sparse belief associated with BCS to a belief that the estimated signals will match the predicted signals given a known DOA change.  This is done by modelling the difference between the expected sparse received signals and the estimated sparse received signals as a Gaussian distribution.  Example test scenarios are provided and comparisons made with the traditional BCS based estimation method.  They show that an improvement in estimation accuracy is possible without a significant increase in computational complexity.

\end{abstract}

\section{Introduction}
Direction of arrival (DOA) estimation is the process of determining which direction a signal impinging on an array has arrived from \cite{vantrees02a}.  Commonly used methods of solving this problem are: MUSIC \cite{Schmidt86,Swindlehurst92} and ESPRIT \cite{Roy89,Gao05}.  However, these methods have two drawbacks:  Firstly, we need some knowledge of the number of signals that are present.  Secondly, evaluation of the covariance matrix is required, thus increasing the computational complexity required to solve the problem.  This covariance matrix is estimated form the signals received by each sensor at different time snapshots. Instead, if we consider the fact that only a few of the potential DOAs will have a signal present then we can consider the problem from the view point of compressive sensing (CS) and work directly with the received signals.

CS theory tells us that when certain conditions are met it is possible to recover
some signals from fewer measurements than used by traditional
methods \cite{Candes06,Donoho06}.  This can be applied to solve the problem of DOA estimation \cite{Mali05,Hyder09,Shen14}.  First split the angular region of interest into $N$ potential DOAs, where signals actually impinge on the array from only $L$ ($L<<N$) of these directions.  The problem can then be formulated as finding the minimum number of DOAs with a signal present that still gives an acceptable approximation of the array output.  Those directions that have the non-zero valued signals are then used as the DOA estimates.  It is also possible to convert this problem into a probabilistic form and solve using a relevance vector machine (RVM) based approach \cite{Ji08,Tipping01,Tipping03}.  It has been shown in the case of static DOA estimation that methods based on this approach offer encouraging results \cite{Carlin13}.

Less work has been done on the problem of estimating a dynamic DOA.  One option is to use particle filters or probability hypothesis density (PHD) filters.  These filters have been used in the areas of DOA estimation and tracking of sources \cite{Ward03,Bala05,Zhong13,Baum14}.

Alternatively, in \cite{Khom10} the authors track a dynamic DOA with a Kalman filter (KF) and narrow the angular region being considered to focus in more closely on the DOA estimate from the previous iteration.  However, this removes the advantage of being able to directly work with the measured array signals and introduces an additional stage of having to reevaluate the steering vector of the array at each iteration of the KF.

Bayesian Kalman filters (BKF) have been used to track dynamic sparse signals \cite{Karseras13}, where the predicted mean of the signals at each iteration is taken as the estimate from the previous iteration and the hyper-parameters (precision) are estimated using BCS, hence the term Bayesian compressed sensing Kalman Filter (BCSKF).  There are still some issues with this method when applied to the problem of DOA estimation of a dynamic far field source using uniform linear arrays (ULAs).  Namely when the DOA approaches the endfire region (i.e. the signal arrives parallel/close to parallel to the array), the estimation accuracy can degrade.  This means that it is possible to initially have an accurate estimate and then not track the changes in DOA properly.

In this paper, to solve this problem, we propose modifying the BCSKF to include information about what the received array signals would be expected to be at a given time snapshot.  This is done by changing the distribution used to model the received signals.  Now instead of assuming a zero-mean Gaussian hierarchial prior we assume that the mean is instead centred at the value that is expected given the previous snapshot's estimate and the expected change in DOA.  As a result we derive a new posterior distribution and marginal likelihood function that can be used to solve the DOA estimation problem by following a similar framework as used for the RVM.  We term this framework the modified RVM, which is used to find an initial estimate of the DOA at the first snapshot and then at each subsequent snapshot to optimise the noise variance estimate as well as the hyperparameters required by the BCSKF.

The remainder of this paper is structured in the following manner:
Section \ref{sec:design} gives details of the proposed estimation method.
This includes the array model being used (\ref{sub:AM}), the modified RVM framework for BCS (\ref{sub:BCS}) and the BCSKF (\ref{sub:BCSKF}).
In Section \ref{sec:sim} an evaluation of the effectiveness of the proposed method is presented and conclusions are
drawn in Section \ref{sec:con}.

\section{Proposed Design Methods}\label{sec:design}
\subsection{Array Model}\label{sub:AM}
\begin{figure}
\begin{center}
   \includegraphics[angle=0,width=0.37\textwidth]{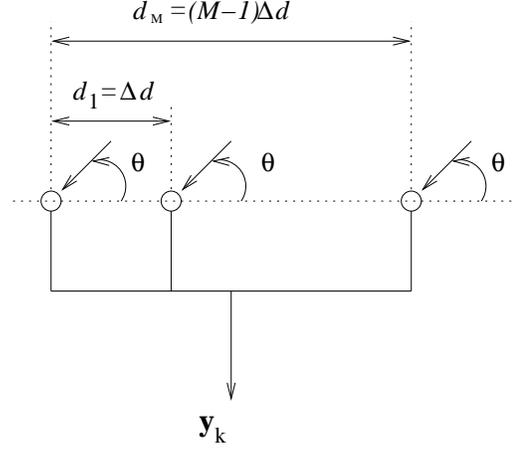}
   \caption{Linear Array structure being considered, consisting of $M$ sensors with a uniform adjacent sensor separation of $\Delta d$.
    \label{fig:AM}}
\end{center}
\end{figure}
A narrowband array structure consisting of $M$ sensors is shown in
Fig. \ref{fig:AM}. The sensors are assumed to be omnidirectional
with identical responses.  A plane-wave signal model is
assumed, i.e. the signal impinges upon the array from the far field at an angle $\theta$ as shown.  In this work we assume that $0^{\circ}\leq\theta\leq180^{\circ}$.
The distance from the first sensor to subsequent sensors is
denoted as $d_{m}$ for $m = 1, 2, \ldots, M$, with $d_{1}=0$, i.e.
the distance from the first sensor to itself.  Note, these values are multiples of a uniform adjacent sensor separation of $\Delta d$.

The steering vector of the array is given by
\begin{equation}
\label{eq:A}
    \textbf{a}(\Omega,\theta)=[1, e^{-j\mu_{2}\Omega
    \cos\theta}, \ldots, e^{-j\mu_{M}\Omega
    \cos\theta}]^{T},
\end{equation}
where $\Omega=\omega T_{s}$ is the normalised frequency with $T_s$
being the sampling period, $\mu_{m}=\frac{d_{m}}{cT_{s}}$ for $m=1,
2, \ldots, M$, and $\{\cdot\}^{T}$ denotes the transpose operation.  Note, the steering vector of an array gives contains information about the array geometry, namely the sensor locations and the delay required for a signal to reach a given sensor.

The output of the array, $\textbf{y}_{k}$, at time snapshot $k$ is then given by
\begin{equation}\label{eq:y}
    \textbf{y}_{k} = \textbf{Ax}_{k} + \textbf{n}_{k},
  \end{equation}
where $\textbf{x}_{k}=[x_{k,1}, x_{k,2}, ..., x_{k,N}]^{T}\in\mathbb{C}^{N\times 1}$ gives the received signals, $\textbf{n}_{k}=[n_{k,1}, n_{k,2}, ..., n_{k,M}]^{T}\in\mathbb{C}^{M\times 1}$ the sensor noise and $\textbf{A}=[\textbf{a}(\Omega,\theta_{1}), \textbf{a}(\Omega,\theta_{2}), ..., \textbf{a}(\Omega,\theta_{N})]\in\mathbb{C}^{M\times N}$ is the matrix containing the steering vectors for each angle of interest.

\subsection{Bayesian Compressed Sensing for DOA Estimation}\label{sub:BCS}
First split the angular range that is being monitored into $N$ potential DOAs.  Each direction can then be considered as having a signal present.  However, only $L<<N$ of the received signals are non-zero valued, with the directions of these $L$ signals giving the actual DOAs.

Now we split (\ref{eq:y}) into real and imaginary components in the following way
\begin{align}\label{eq:y split}
    \tilde{\textbf{y}}_{k} &= \tilde{\textbf{A}}\tilde{\textbf{x}}_{k}+\tilde{\textbf{n}}_{k} \\ \nonumber
    \left[
      \begin{array}{c}
        \mathcal{R}(\textbf{y}_{k}) \\
        \mathcal{I}(\textbf{y}_{k}) \\
      \end{array}
    \right]
     &= \left[
           \begin{array}{cc}
             \mathcal{R}(\textbf{A}) & -\mathcal{I}(\textbf{A}) \\
             \mathcal{I}(\textbf{A}) & \mathcal{R}(\textbf{A}) \\
           \end{array}
         \right] \left[
                  \begin{array}{c}
                    \mathcal{R}(\textbf{x}_{k}) \\
                    \mathcal{I}(\textbf{x}_{k}) \\
                  \end{array}
                \right] + \left[
                         \begin{array}{c}
                           \mathcal{R}(\textbf{n}_{k}) \\
                           \mathcal{I}(\textbf{n}_{k}) \\
                         \end{array}
                       \right],\vspace*{.75cm}
  \end{align}
where $\mathcal{R}(\cdot)$ and $\mathcal{I}(\cdot)$ give the real and imaginary components respectively.  Note, the difference between $\textbf{y}_{k}$ and $\tilde{\textbf{y}}_{k}$ is that $\textbf{y}_{k}$ has been split into its real and imaginary components in $\tilde{\textbf{y}}_{k}$.  As a result the dimensions of $\tilde{\textbf{y}}_{k}$ are larger than for $\textbf{y}_{k}$, but we are now only considering real valued data.  Similar is true when comparing $\textbf{A}$ and $\tilde{\textbf{A}}$, and $\textbf{x}_{k}$ and $\tilde{\textbf{x}}_{k}$ and $\textbf{n}_{k}$ and $\tilde{\textbf{n}}_{k}$.

The aim is to now find a solution for $\tilde{\textbf{x}}_{k}$ that gives the minimum $l_{0}$ norm, i.e. the minimum number of non-zero valued signals.   This is done by evaluating the following
\begin{equation}\label{eq:xest1}
    \tilde{\textbf{x}}_{k,opt} = \max\mathcal{P}(\tilde{\textbf{x}}_{k},\sigma^{2},\textbf{p}|\tilde{\textbf{y}},\textbf{x}_{e}),
  \end{equation}
  where $\sigma$ is the variance of the Gaussian noise $\textbf{n}$, $\textbf{p}=[p_{1}, p_{2}, ..., p_{2N}]^{T}]$ contains the hyperparameters that are to be estimated and $\textbf{x}_{e}$ holds the expected values of $\tilde{\textbf{x}}_{k}$.

To do this first obtain the following from \eqref{eq:y}:
\begin{equation}\label{eq:y dist}
  \mathcal{P}(\tilde{\textbf{y}}_{k}|\tilde{\textbf{x}}_{k},\sigma^{2}) = (2\pi\sigma^{2})^{-M}\exp\Big\{-\frac{1}{2\sigma^{2}}||\tilde{\textbf{y}}_{k} - \tilde{\textbf{A}}\tilde{\textbf{x}}_{k}||_{2}^{2}\Big\}.
\end{equation}
Now exert a belief about the values of $\tilde{\textbf{x}}_{k}$ that are expected by enforcing
\begin{eqnarray}\label{eq:x dist}
  \mathcal{P}(\tilde{\textbf{x}}_{k}|\textbf{p},\textbf{x}_{e}) &= & (2\pi)^{-N}|\textbf{P}|^{1/2}\\ \nonumber&\times&\exp\Big\{-\frac{1}{2}(\tilde{\textbf{x}}_{k} - \textbf{x}_{e})\textbf{P}(\tilde{\textbf{x}}_{k} - \textbf{x}_{e})^{T}\Big\},
\end{eqnarray}
where $|\textbf{P}|$ indicates the determinant of $\textbf{P}$, where $\textbf{P}=\text{diag}(\textbf{p})$.

It is also necessary to define the hyperparameters over $\textbf{p}$ and $\sigma^{2}$.  There are various possibilities for the structuring of the priors on $\textbf{p}$, which represent mixing parameters in a scale mixture of normals representation of the marginal distribution of $\textbf{x}_{k}$, which will here be in the Student-t family, see e.g. \cite{And74}.  One possibility would be to treat the complex components of $\textbf{x}_{k}$ as complex Student-t distributed, as detailed in \cite{Wolfe02,Wolfe04}.  Here though we treat the real and imaginary components of $\textbf{x}_{k}$ as independent Student-t distributed random variables, and hence have independent Gamma priors for the mixing variables $p_{n}$ over all real and imaginary components of $\textbf{x}_{k}$:
 \begin{equation}\label{eq:bcs8}
  \mathcal{P}(\textbf{p})=\prod_{n=1}^{2N}G(p_{n}|\beta_{1},\beta_{2}).
\end{equation}
A Gamma prior can also be used for $\sigma^{2}$
\begin{equation}\label{eq:bcs9}
  \mathcal{P}(\sigma^{2})=G(\sigma^{-2}|\beta_{3},\beta_{4}),
\end{equation}
where $\beta_{1},\beta_{2},\beta_{3}$ and $\beta_{4}$ are scale and shape priors.  Note, when $\textbf{x}_{e} = [0, 0, ..., 0]^{T}$ then (\ref{eq:x dist}) reverts to the traditional hierarchial prior used in BCS \cite{Ji08}.

We know that
\begin{equation}\label{eq:BCS1}
  \mathcal{P}(\tilde{\textbf{x}}_{k},\sigma^{2},\textbf{p}|\tilde{\textbf{y}}_{k},\textbf{x}_{e})=\mathcal{P}(\tilde{\textbf{x}}_{k}|\tilde{\textbf{y}}_{k},\sigma^{2},\textbf{p},\textbf{x}_{e})\mathcal{P}(\textbf{p},\sigma^{2}|\tilde{\textbf{y}}_{k})
\end{equation}
and \footnote{See Appendix A}
\begin{eqnarray}\label{eq:post}
  \mathcal{P}(\tilde{\textbf{x}}_{k}|\tilde{\textbf{y}}_{k},\textbf{p},\sigma^{2},\textbf{x}_{e}) &=& \frac{\mathcal{P}(\tilde{\textbf{y}}_{k}|\tilde{\textbf{x}}_{k},\sigma^{2})\mathcal{P}(\tilde{\textbf{x}}_{k}|\textbf{p},\textbf{x}_{e})}{\mathcal{P}(\tilde{\textbf{y}}_{k}|\textbf{p},\sigma^{2},\textbf{x}_{e})}\\ \nonumber
  &=&(2\pi)^{-N}|\boldsymbol\Sigma|^{-1/2}\\ \nonumber&\times&\exp\Bigg\{ -\frac{1}{2}(\tilde{\textbf{x}}_{k}-\boldsymbol\mu)^{T}\boldsymbol\Sigma^{-1}(\tilde{\textbf{x}}_{k}-\boldsymbol\mu)  \Bigg\},
\end{eqnarray}
where the covariance matrix is given by
\begin{equation}\label{eq:SIGMA}
  \boldsymbol\Sigma = (\sigma^{-2}\tilde{\textbf{A}}^{T}\tilde{\textbf{A}}+\textbf{P})^{-1},
\end{equation}
and the mean given by
\begin{equation}\label{eq:mu}
  \boldsymbol\mu = \boldsymbol\Sigma(\sigma^{-2}\tilde{\textbf{A}}^{T}\tilde{\textbf{y}}_{k}+\textbf{P}\textbf{x}_{e}).
\end{equation}
Note, the maximum of (\ref{eq:post}) is the posterior mean $\boldsymbol\mu$.

Similarly to \cite{Tipping01}, the probability $\mathcal{P}(\sigma^{2},\textbf{p}|\tilde{\textbf{y}}_{k})$ can be represented in the following form:
\begin{equation}\label{eq:BCS2}
  \mathcal{P}(\sigma^{2},\textbf{p}|\tilde{\textbf{y}}_{k})\approx\mathcal{P}(\tilde{\textbf{y}}_{k}|\textbf{p},\sigma^{2},\textbf{x}_{e})\mathcal{P}(\textbf{p})\mathcal{P}(\sigma^{2})
\end{equation}
and the second two terms on the right of (\ref{eq:BCS2}) become constant with uniform scale priors, then maximising $\mathcal{P}(\sigma^{2},\textbf{p}|\tilde{\textbf{y}}_{k})$ is roughly equivalent to maximising $\mathcal{P}(\tilde{\textbf{y}}_{k}|\textbf{p},\sigma^{2},\textbf{x}_{e})$.  This can be achieved by a type 2 maximisation of its logarithm, which is given by \footnote{See Appendix B}
\begin{eqnarray}\label{eq:log}
  \mathcal{L}(\textbf{p},\sigma^{2}) &=& \log\Bigg\{ (2\pi\sigma^{2})^{-M}|\boldsymbol\Sigma|^{\frac{1}{2}}|\textbf{P}|^{\frac{1}{2}}\exp\Big(-\frac{1}{2} \\ \nonumber
   &\times&(\tilde{\textbf{y}}_{k}^{T}\textbf{B}\tilde{\textbf{y}}_{k}+\textbf{x}_{e}^{T}\textbf{C}\textbf{x}_{e}-2\sigma^{2}\tilde{\textbf{y}}_{k}^{T}\tilde{\textbf{A}}\boldsymbol\Sigma\textbf{Px}_{e})\Big) \Bigg\} \\ \nonumber
   &=& -\frac{1}{2}\Big(2M\log(2\pi)+2M\log\sigma^{2}-\log|\boldsymbol\Sigma| - \\ \nonumber
   &&\log|\textbf{P}|+\sigma^{-2}||\tilde{\textbf{y}}_{k}-\tilde{\textbf{A}}\boldsymbol\mu||^{2}_{2} + \boldsymbol\mu^{T}\textbf{P}\boldsymbol\mu \\ \nonumber
   &&+\textbf{x}_{e}^{T}\textbf{Px}_{e}-\textbf{x}_{e}^{T}\textbf{P}\boldsymbol\mu\Big),
\end{eqnarray}
where $\textbf{B}=(\sigma^{2}\textbf{I}+\tilde{\textbf{A}}\textbf{P}^{-1}\tilde{\textbf{A}}^{T})^{-1}$ and $\textbf{C}=\textbf{P}-\textbf{P}^{T}\boldsymbol\Sigma\textbf{P}$.

To do this (\ref{eq:log}) is differentiated with respect to $p_{n}$ and $\sigma^{-2}$ to obtain the update expressions \footnote{See Appendix C}
\begin{equation}\label{eq:pnew}
  p_{n}^{new}=\frac{\gamma_{n}}{\mu_{n}^{2}+x_{e,n}^{2}-x_{e,n}\mu_{n}},
\end{equation}
where $\gamma_{n}=1-p_{n}\Sigma_{nn}$, $\Sigma_{nn}$ is the $n^{th}$ diagonal element of $\boldsymbol\Sigma$ and
\begin{equation}\label{eq:sigmanew}
  \sigma_{new}^{2}=\frac{||\tilde{\textbf{y}}_{k}-\tilde{\textbf{A}}\boldsymbol\mu||_{2}^{2}}{2M-\sum\limits_{n}\gamma_{n}}.
\end{equation}
The maximisation is then achieved by iteratively maximising (\ref{eq:SIGMA}) and (\ref{eq:mu}) and (\ref{eq:pnew}) and (\ref{eq:sigmanew}) until a convergence criterion is met \cite{Ji08,Tipping01}.  Note that when $\textbf{x}_{e} = [0, 0, ..., 0]^{T}$ the update expressions match that of the traditional RVM.  The final estimate of the received signals is then given by
\begin{equation}\label{eq:xopt}
  \tilde{\textbf{x}}_{k,opt}=\Big(\frac{\tilde{\textbf{A}}^{T}\tilde{\textbf{A}}}{\sigma_{opt}^{2}}+\textbf{P}_{opt}\Big)^{-1}\Big( \frac{\tilde{\textbf{A}}^{T}\tilde{\textbf{y}}_{k}}{\sigma_{opt}^{2}}+\textbf{P}_{opt}\textbf{x}_{e} \Big)
\end{equation}
where $\sigma^{2}_{opt}$ and $\textbf{P}_{opt}=\text{diag}([p_{opt,1}, p_{opt,2}, ..., p_{opt,2N}]^{T})$ are the result of optimising the noise estimate and hyperparameters respectively.

The final estimated signals are then given by
\begin{equation}\label{eq:xfinal}
  x_{k,opt,n}=\tilde{x}_{k,opt,n}+j\tilde{x}_{k,opt,N+n}.
\end{equation}
Thresholding can then be applied to keep the $\tilde{L}$ most significant signals as in \cite{Carlin13}.  To do this find the total energy content of the estimated received signals and then sort them.  A threshold value, $\eta$, is then defined as a percentage of the energy content that is to be retained.  Starting with the most significant estimated signal, the estimated signals are summed until the threshold is reached and the remaining signals set to be equal to 0.  The remaining non-zero valued signals then give the DOA estimates and $\tilde{L}$ as an estimate of the number of far field signals impinging on the array.
\subsection{Bayesian Compressed Sensing Kalman Filter}\label{sub:BCSKF}
In order to track the changes in the DOA estimates at each time snapshot the BCS based DOA estimation procedure detailed above is combined with a BKF, giving a BCSKF for DOA estimation.  Here, the signal model described above is again used along with the prediction
\begin{align}\nonumber
    \tilde{\textbf{x}}_{k|k-1} &= \tilde{\textbf{x}}_{{k-1}|k-1}+\boldsymbol\Delta\textbf{x} & \boldsymbol\Sigma_{k|k-1} &= \boldsymbol\Sigma_{k-1} + \textbf{P}^{-1}_{k} \\
    \tilde{\textbf{y}}_{k|k-1} &= \tilde{\textbf{A}}\tilde{\textbf{x}}_{k|k-1}& \tilde{\textbf{y}}_{e,k} &= \tilde{\textbf{y}}_{k}-\tilde{\textbf{y}}_{k|k-1}
  \end{align}
and update steps
  \begin{align}\nonumber
  \tilde{\textbf{x}}_{k} = \tilde{\textbf{x}}_{k|k-1} + \textbf{K}_{k}\tilde{\textbf{y}}_{e,k}&\;\;\; \boldsymbol\Sigma_{k|k}=(\textbf{I}-\textbf{K}_{k}\tilde{\textbf{A}})\boldsymbol\Sigma_{k|k-1}\\
   \textbf{K}_{k}=\boldsymbol\Sigma_{k|k-1}\tilde{\textbf{A}}^{T}&(\sigma^{2}\textbf{I}+\tilde{\textbf{A}}\boldsymbol\Sigma_{k|k-1}\tilde{\textbf{A}}^{T})^{-1}
  \end{align}
of the BKF.  Here, $k|k-1$ indicates prediction at time instance $k$ given the previous measurements and $\boldsymbol\Delta\textbf{x}$ is determined by the expected DOA change.  For example, if we sample the angular range every $1^{\circ}$ and the the DOA increases by $2^{\circ}$ then then $\boldsymbol\Delta\textbf{x}$ will be selected to increase the index of the non-zero valued entries in $\tilde{\textbf{x}}_{k-1|k-1}$ by two to give the index of the non-zero valued entries in $\tilde{\textbf{x}}_{k|k-1}$.  In this work we have assumed that there will be a constant change in the DOA.

At each time snapshot it is necessary to estimate the noise variance and hyperparameters in order to evaluate the prediction and update steps of the BCSKF.  This is done by considering the log likelihood function given by
\begin{eqnarray}
  \mathcal{L}(\textbf{p}_k,\sigma^{2}) &=& -\frac{1}{2}\Big(2M\log(2\pi)+2M\log\sigma^{2}-\\ \nonumber&&\log|\boldsymbol\Sigma| -
   \log|\textbf{P}|+\sigma^{-2}||\tilde{\textbf{y}}_{e,k}-\tilde{\textbf{A}}\boldsymbol\mu||^{2}_{2}\\ \nonumber
   && + \boldsymbol\mu^{T}\textbf{P}\boldsymbol\mu +\tilde{\textbf{x}}_{k|k-1}^{T}\textbf{P}\tilde{\textbf{x}}_{k|k-1}-\tilde{\textbf{x}}_{k|k-1}^{T}\textbf{P}\boldsymbol\mu\Big),
\end{eqnarray}
which can be optimised by following the procedure described in Section \ref{sub:BCS}.  Here we have used the Kalman filter prediction $\tilde{\textbf{x}}_{k|k-1}$ as the expected estimate values $\textbf{x}_{e}$.

It is worth noting that the continued accuracy of the proposed BCSKF relies on the accuracy of the initial estimate and the parameter values selected.  If the initial estimate (made using the framework described in Section \ref{sub:BCS} and $\textbf{x}_{e}=[0, 0, ..., 0]^{T}$) of the received signals is accurate and sparse, then the priors that are enforced will ensure this continues to be the case.  However, an inaccurate initial DOA estimate or poorly matched expected DOA change can lead to the introduction of inaccuracies in subsequent estimates.  Similarly, if the initial estimate of the received signals is not sparse then subsequent estimates are likely to not be sparse.  As a result, care should be taken when choosing the initial parameter values and determining the likely DOA change.
\section{Performance Evaluation}\label{sec:sim}
In this section a comparison in performance of the traditional RVM and the proposed modified RVM will be made.  Three example scenarios will be considered.  One where the initial DOA starts outside of the endfire region and then moves into it, one where the DOA remains out of the endfire region and finally one where the initial DOAs and signal values are randomly generated.  All of the examples are are implemented in Matlab on a computer with
an Intel Xeon CPU E3-1271 (3.60GHz) and 16GB of RAM.

The performance of each method will be measured using the root mean square error ($RMSE$) in DOA estimate.  This is given by
\begin{equation}\label{eq:RMSE}
  RMSE = \sqrt{\frac{\sum\limits_{q=1}^{Q}|\theta-\hat{\theta}|^{2}}{Q}},
\end{equation}
where $\theta$ is the actual DOA, $\hat{\theta}$ is the estimated DOA and $Q$ is the number of Monte Carlo simulations carried out, with $Q=100$ being used in each case.

The array structure being considered is a 20 sensor ULA with an adjacent sensor separation of $\frac{\lambda}{2}$.  We assume the actual noise variance is given by $\sigma_{n}^{2}=0.4$ and an initial estimate of the noise variance of $\sigma_{init}^{2}=0.1$ used when initialising the RVM and proposed modified RVM.

\subsection{Endfire Region}
\begin{figure}
\begin{center}
   \includegraphics[angle=0,width=0.4\textwidth]{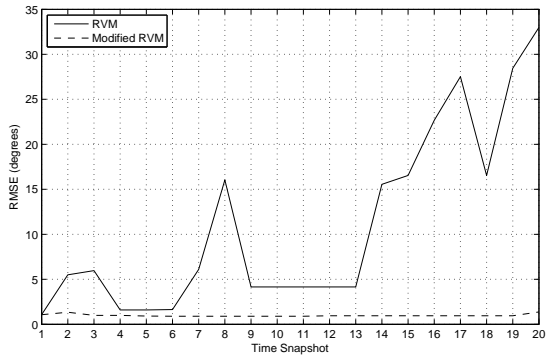}
   \caption{$RMSE$ values for the endfire region example.
    \label{fig:end}}
\end{center}
\end{figure}

For this example the initial DOA of the signal is $\theta=20^{\circ}$, which then decreases by $1^{\circ}$ at each time snapshot.  The signal value at each snapshot is set to be 1.  Table \ref{tb:end} summarises the performance of the two methods for this example.  Here we can see that there is in an improved accuracy in the DOA estimates, as the mean $RMSE$ has decreased for the proposed modified RVM method.  This is also supported by the overall $RMSE$ values as illustrated in Fig. \ref{fig:end}.  It is also worth noting that the mean computation times show that this improvement has not been at the expense of a significant increase in computational complexity.

\begin{table}
\caption{Performance summary for the endfire region example.} \centering
\begin{tabular}{|c|c|c|}
  \hline
   & Mean $RMSE$  & Mean Computation  \\
  Method&(degrees) &Time (seconds) \\
  \hline
  RVM & 11.03 & 0.35 \\
  \hline
  Modified RVM & 0.98 & 0.55 \\
  \hline
\end{tabular}
\label{tb:end}
\end{table}

\subsection{Non-Endfire Region}
\begin{figure}
\begin{center}
   \includegraphics[angle=0,width=0.4\textwidth]{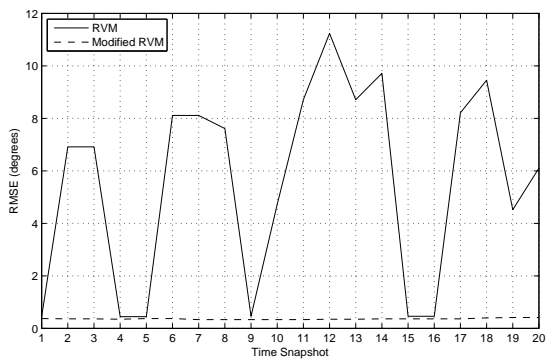}
   \caption{$RMSE$ values for the non-endfire region example.
    \label{fig:non}}
\end{center}
\end{figure}

In this instance the initial DOA is $\theta=100^{\circ}$ with the DOA increasing by $1^{\circ}$ at each time snapshot, with the signal value remaining constant at -1.  The performance of the two methods is summarised in Table \ref{tb:non}, with the $RMSE$ values illustrated in Fig. \ref{fig:non}.  Again this illustrates the improved performance offered by the modified RVM has not been at the expense of a significant increase in computation time.
\begin{table}
\caption{Performance summary for the non-endfire region example.} \centering
\begin{tabular}{|c|c|c|}
  \hline
   & Mean $RMSE$  & Mean Computation  \\
  Method&(degrees) &Time (seconds) \\
  \hline
  RVM & 5.59 & 0.33 \\
  \hline
  Modified RVM & 0.36 & 0.41 \\
  \hline
\end{tabular}
\label{tb:non}
\end{table}

\subsection{Random Initial DOA}
\begin{figure}
\begin{center}
   \includegraphics[angle=0,width=0.4\textwidth]{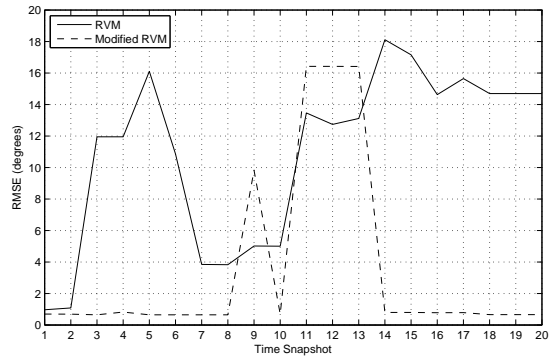}
   \caption{$RMSE$ values for the random initial DOA example.
    \label{fig:ran}}
\end{center}
\end{figure}

Finally, we consider the case where the initial DOA is randomly chosen from the entire angular range and increased by $1^{\circ}$ at each time snapshot.  The signal value is randomly selected as $\pm1$ for each simulation and remains constant as the DOA changes.  As for the previous two examples Table. \ref{tb:ran} and Fig. \ref{fig:ran} indicate that the proposed modified RVM has obtained an improved accuracy without a significant increase in computational complexity.
\begin{table}
\caption{Performance summary for the random initial DOA example.} \centering
\begin{tabular}{|c|c|c|}
  \hline
   & Mean $RMSE$  & Mean Computation  \\
  Method&(degrees) &Time (seconds) \\
  \hline
  RVM & 10.98 & 0.34 \\
  \hline
  Modified RVM & 3.52 & 0.43 \\
  \hline
\end{tabular}
\label{tb:ran}
\end{table}

\section{Conclusions}\label{sec:con}
This paper proposes a BCSKF to estimate  the DOA of a single signal impinging on a ULA from the far field.  A new posterior distribution and marginal likelihood has been found and unlike traditional BCS the expected values of the estimates are accounted for.  This is done to combat the problem of inaccurate DOA estimates when the actual DOA approaches the endfire region of the angular range.  Then a similar optimisation framework to what is used in the RVM is applied to find the optimal hyperparameters and noise variance estimate, which are then used to estimate the received array signals.  Example test scenarios have shown the proposed modified RVM is more accurate in not only the endfire region, but also in the entire angular region as a whole.  This is also without a significant increase in computational complexity.

\section*{Appendix}
\subsection{Derivation of Posterior Distribution}
From Bayes' rule we know that
\begin{eqnarray}\label{eq:Bayes}
  \mathcal{P}(\tilde{\textbf{x}}_{k}|\tilde{\textbf{y}}_{k},\textbf{p},\sigma^{2},\textbf{x}_{e})\mathcal{P}(\tilde{\textbf{y}}_{k}|\textbf{p},\sigma^{2},\textbf{x}_{e})&=&\\ \nonumber
  \mathcal{P}(\tilde{\textbf{y}}_{k}|\tilde{\textbf{x}}_{k},\sigma^{2})\mathcal{P}(\tilde{\textbf{x}}_{k}|\textbf{p},\textbf{x}_{e}),
\end{eqnarray}
where $\mathcal{P}(\tilde{\textbf{y}}_{k}|\tilde{\textbf{x}}_{k},\sigma^{2})$ and $\mathcal{P}(\tilde{\textbf{x}}_{k}|\textbf{p},\textbf{x}_{e})$ are known from (\ref{eq:y dist}) and (\ref{eq:x dist}), respectively.

Now following the method suggested in \cite{Tipping01} carry out the multiplication on the right hand side, collect terms in $\tilde{\textbf{x}}_{k}$ in the exponential and complete the square.
\begin{eqnarray}
  && -\frac{1}{2}\Big[\sigma^{-2}(\tilde{\textbf{y}}_{k} - \tilde{\textbf{A}}\tilde{\textbf{x}}_{k})^{T}(\tilde{\textbf{y}}_{k} - \tilde{\textbf{A}}\tilde{\textbf{x}}_{k})+\\ \nonumber &&\;\;\;\;\;\;\;\;\;\;\;\;\;\;\;\;\;\;(\tilde{\textbf{x}}_{k} - \textbf{x}_{e})^{T}\textbf{P}(\tilde{\textbf{x}}_{k} - \textbf{x}_{e})\Big] \\ \nonumber
  &=& -\frac{1}{2}\Big[ \sigma^{-2}\tilde{\textbf{y}}_{k}^{T}\tilde{\textbf{y}}_{k} - \sigma^{-2}\tilde{\textbf{y}}_{k}^{T}\tilde{\textbf{A}}\tilde{\textbf{x}}_{k} - \sigma^{-2}\tilde{\textbf{x}}_{k}^{T}\tilde{\textbf{A}}^{T}\tilde{\textbf{y}}_{k} +\\ \nonumber && \sigma^{-2}\tilde{\textbf{x}}_{k}^{T}\tilde{\textbf{A}}^{T}\tilde{\textbf{A}}\tilde{\textbf{x}}_{k} + \tilde{\textbf{x}}_{k}^{T}\textbf{P}\tilde{\textbf{x}}_{k} - \tilde{\textbf{x}}_{k}^{T}\textbf{Px}_{e} - \textbf{x}_{e}^{T}\textbf{P}\tilde{\textbf{x}}_{k}\\ \nonumber && \;\;\;\;\;\;\;\;\;\;\;\;\;\;\;\;\;\;\;\;\;\;\;\;\;\;\;\;\;\;\;\;\;\;\;\; + \textbf{x}_{e}^{T}\textbf{Px}_{e} \Big] \\ \nonumber
   &=&  -\frac{1}{2}\Big[ (\tilde{\textbf{x}}_{k}-\boldsymbol\mu)^{T}\boldsymbol\Sigma^{-1}(\tilde{\textbf{x}}_{k}-\boldsymbol\mu) - \boldsymbol\mu^{T}\boldsymbol\Sigma^{-1}\boldsymbol\mu + \\ \nonumber && \;\;\;\;\;\;\;\;\;\;\;\;\;\;\;\;\;\;\;\;\; \sigma^{-2}\tilde{\textbf{y}}_{k}^{T}\tilde{\textbf{y}}_{k} + \textbf{x}_{e}^{T}\textbf{Px}_{e} \Big]
\end{eqnarray}
where $\boldsymbol\Sigma$ and $\boldsymbol\mu$ are given by (\ref{eq:SIGMA}) and (\ref{eq:mu}), respectively. This then gives the posterior distribution as (\ref{eq:post}), with the remaining exponential terms
\begin{equation}\label{eq:remain1}
  -\frac{1}{2}\Bigg[ \sigma^{-2}\tilde{\textbf{y}}_{k}^{T}\tilde{\textbf{y}}_{k} + \textbf{x}_{e}^{T}\textbf{Px}_{e} - \boldsymbol\mu^{T}\boldsymbol\Sigma^{-1}\boldsymbol\mu  \Bigg].
\end{equation}
\subsection{Derivation of Marginal Likelihood}
From (\ref{eq:Bayes}) we know that
\begin{equation}\label{eq:Bayes2}
  \mathcal{P}(\tilde{\textbf{y}}_{k}|\textbf{p},\sigma^{2},\textbf{x}_{e})=\frac{\mathcal{P}(\tilde{\textbf{y}}_{k}|\tilde{\textbf{x}}_{k},\sigma^{2}),\mathcal{P}(\tilde{\textbf{x}}_{k}|\textbf{p},\textbf{x}_{e})}{\mathcal{P}(\tilde{\textbf{x}}_{k}|\tilde{\textbf{y}}_{k},\textbf{p},\sigma^{2},\textbf{x}_{e})},
\end{equation}
meaning the term in the exponential will be (\ref{eq:remain1}) where
\begin{eqnarray}\label{eq:musigmu}
  \boldsymbol\mu^{T}\boldsymbol\Sigma^{-1}\boldsymbol\mu&=&(\sigma^{-2}\tilde{\textbf{A}}^{T}\tilde{\textbf{y}}_{k}+\textbf{P}\tilde{\textbf{x}}_{e})^{T}\boldsymbol\Sigma^{T}\boldsymbol\Sigma^{-1}\\ \nonumber &\times&\boldsymbol\Sigma(\sigma^{-2}\tilde{\textbf{A}}^{T}\tilde{\textbf{y}}_{k}+\textbf{Px}_{e})\\ \nonumber
  &=&(\sigma^{-2}\tilde{\textbf{A}}^{T}\tilde{\textbf{y}}_{k}+\textbf{Px}_{e})^{T}
  (\sigma^{-2}\boldsymbol\Sigma\tilde{\textbf{A}}^{T}\tilde{\textbf{y}}_{k}+\boldsymbol\Sigma\textbf{Px}_{e})\\ \nonumber &=&\sigma^{-4}\tilde{\textbf{y}}_{k}^{T}\tilde{\textbf{A}}\boldsymbol\Sigma\tilde{\textbf{A}}^{T}\tilde{\textbf{y}}_{k} + \sigma^{-2}\tilde{\textbf{y}}_{k}^{T}\tilde{\textbf{A}}\boldsymbol\Sigma\textbf{Px}_{e} + \\ \nonumber&& \sigma^{-2}\textbf{x}_{e}^{T}\textbf{P}^{T}\tilde{\textbf{A}}^{T}\tilde{\textbf{y}}_{k} + \textbf{x}_{e}^{T}\textbf{P}^{T}\boldsymbol\Sigma\textbf{Px}_{e}.
\end{eqnarray}
Therefore the exponential term is given by
\begin{eqnarray}\label{eq:exp}
 && -\frac{1}{2}\Bigg[\sigma^{-2}\tilde{\textbf{y}}_{k}^{T}\tilde{\textbf{y}}_{k} + \textbf{x}_{e}^{T}\textbf{Px}_{e} - \sigma^{-4}\tilde{\textbf{y}}_{k}^{T}\tilde{\textbf{A}}\boldsymbol\Sigma\tilde{\textbf{A}}^{T}\tilde{\textbf{y}}_{k} - \\ \nonumber && \sigma^{-2}\tilde{\textbf{y}}_{k}^{T}\tilde{\textbf{A}}\boldsymbol\Sigma\textbf{Px}_{e} - \sigma^{-2}\textbf{x}_{e}^{T}\textbf{P}^{T}\tilde{\textbf{A}}^{T}\tilde{\textbf{y}}_{k} - \\ \nonumber &&\;\;\;\;\;\;\;\;\;\;\;\;\;\;\;\;\;\;\; \textbf{x}_{e}^{T}\textbf{P}^{T}\boldsymbol\Sigma\textbf{Px}_{e}   \Bigg] \\ \nonumber
   &=&  -\frac{1}{2}\Bigg[\tilde{\textbf{y}}_{k}^{T}[\sigma^{-2}-\sigma^{-4}\tilde{\textbf{A}}\boldsymbol\Sigma\tilde{\textbf{A}}^{T}]
   \tilde{\textbf{y}}_{k} + \textbf{x}_{e}^{T}[\textbf{P}-\textbf{P}^{T}\boldsymbol\Sigma\textbf{P}]\textbf{x}_{e}\\ \nonumber && -   \sigma^{-2}\tilde{\textbf{y}}_{k}^{T}\tilde{\textbf{A}}\boldsymbol\Sigma\textbf{Px}_{e} - \sigma^{-2}\textbf{x}_{e}^{T}\textbf{P}^{T}\tilde{\textbf{A}}^{T}\tilde{\textbf{y}}_{k}  \Bigg]
\end{eqnarray}

The term outside of the exponential is given by
\begin{equation}\label{eq:nonexp}
  \frac{(2\pi\sigma^{2})^{-M}(2\pi)^{-N}|\textbf{P}|^{1/2}}{(2\pi)^{-N}|\boldsymbol\Sigma|^{-\frac{1}{2}}}
=(2\pi\sigma^{2})^{-M}|\boldsymbol\Sigma|^{\frac{1}{2}}|\textbf{P}|^{\frac{1}{2}}.
\end{equation}
This gives the marginal likelihood as
\begin{eqnarray}\label{eq:marg}
  \mathcal{P}(\tilde{\textbf{y}}_{k}|\textbf{p},\sigma^{2},\textbf{x}_{e})&=&(2\pi\sigma^{2})^{-M}|\boldsymbol\Sigma|^{\frac{1}{2}}|\textbf{P}|^{\frac{1}{2}}\\ \nonumber&\times&\exp\Big\{ -\frac{1}{2}[\tilde{\textbf{y}}_{k}^{T}\textbf{B}\tilde{\textbf{y}}_{k}+\textbf{x}_{e}^{T}\textbf{C}\tilde{\textbf{x}}_{e}-\\ \nonumber &&\;\;\;\;\;\;\;\;\;\;\;\;\;\;2\sigma^{2}\tilde{\textbf{y}}_{k}^{T}\tilde{\textbf{A}}\boldsymbol\Sigma\textbf{Px}_{e}]\Big\},
\end{eqnarray}
where $\textbf{B}$ and $\textbf{C}$ are defined as in Section \ref{sub:BCS}.  The log likelihood is then given by
\begin{eqnarray}\label{eq:log2}
  \mathcal{L}(\textbf{p},\sigma^{2}) &=& \log\Big\{ (2\pi\sigma^{2})^{-M}|\boldsymbol\Sigma|^{\frac{1}{2}}|\textbf{P}|^{\frac{1}{2}}\\ \nonumber&\times&\exp\Big\{ -\frac{1}{2}[\tilde{\textbf{y}}_{k}^{T}\textbf{B}\tilde{\textbf{y}}_{k}+\textbf{x}_{e}^{T}\textbf{C}\tilde{\textbf{x}}_{e}-\\ \nonumber &&\;\;\;\;\;\;\;\;\;\;\;\;\;\;2\sigma^{2}\tilde{\textbf{y}}_{k}^{T}\tilde{\textbf{A}}\boldsymbol\Sigma\textbf{Px}_{e}]\Big\} \Big\} \\ \nonumber
   &=& -M\log(2\pi) -M\log\sigma^{2} +\frac{1}{2}\log|\boldsymbol\Sigma| +\\ \nonumber &&\frac{1}{2}\log|\textbf{P}| -\frac{1}{2}[\tilde{\textbf{y}}_{k}^{T}\textbf{B}\tilde{\textbf{y}}_{k}+\textbf{x}_{e}^{T}\textbf{C}\tilde{\textbf{x}}_{e}-\\ \nonumber &&\;\;\;\;\;\;\;\;\;\;\;\;\;\;\;\;\;\;\;\;\;\;\;2\sigma^{2}\tilde{\textbf{y}}_{k}^{T}\tilde{\textbf{A}}\boldsymbol\Sigma\textbf{Px}_{e}].
\end{eqnarray}

Using the Woodbury matrix inversion identity we have
\begin{equation}\label{eq:invB}
  \textbf{B} = \sigma^{-2}\textbf{I}-\sigma^{-2}\tilde{\textbf{A}}(\textbf{P}+\sigma^{-2}\tilde{\textbf{A}}^{T}\tilde{\textbf{A}})^{-1}\tilde{\textbf{A}}^{T}\sigma^{-2},
\end{equation}
which means we have
\begin{eqnarray}
  \tilde{\textbf{y}}_{k}^{T}\textbf{B}\tilde{\textbf{y}}_{k} &=& \tilde{\textbf{y}}_{k}^{T}\sigma^{-2}\tilde{\textbf{y}}_{k} - \tilde{\textbf{y}}_{k}^{T}(\sigma^{-2}\textbf{I}-\sigma^{-2}\tilde{\textbf{A}}\\ \nonumber &\times&(\textbf{P}+\sigma^{-2}\tilde{\textbf{A}}^{T}\tilde{\textbf{A}})^{-1}\tilde{\textbf{A}}^{T}\sigma^{-2})\tilde{\textbf{y}}_{k} \\ \nonumber
   &=&  \tilde{\textbf{y}}_{k}^{T}\sigma^{-2}\tilde{\textbf{y}}_{k} - \tilde{\textbf{y}}_{k}^{T}\sigma^{-2}\tilde{\textbf{A}}\boldsymbol\Sigma\tilde{\textbf{A}}^{T}\sigma^{-2}\tilde{\textbf{y}}_{k} \\ \nonumber
   &=& \sigma^{-2}\tilde{\textbf{y}}_{k}^{T}(\tilde{\textbf{y}}_{k}-\tilde{\textbf{A}}\boldsymbol\mu)+\sigma^{-2}\tilde{\textbf{y}}_{k}^{T}\tilde{\textbf{A}}\boldsymbol\Sigma\textbf{Px}_{e} \\ \nonumber
   &=&  \sigma^{-2}||\tilde{\textbf{y}}_{k}^{T}-\tilde{\textbf{A}}\boldsymbol\mu||_{2}^{2}+\boldsymbol\mu^{T}\textbf{P}\boldsymbol\mu+\sigma^{-2}\tilde{\textbf{y}}_{k}^{T}\tilde{\textbf{A}}\boldsymbol\Sigma\textbf{Px}_{e}.
\end{eqnarray}
Also, we know that $\textbf{P}^{T}=\textbf{P}$ as $\textbf{P}$ is a real valued diagonal matrix.  This means
\begin{eqnarray}
  \textbf{x}_{e}^{T}\textbf{Cx}_{e} &=& \textbf{x}_{e}^{T}[\textbf{P}-\textbf{P}\boldsymbol\Sigma\textbf{P}]\textbf{x}_{e} \\ \nonumber
   &=& \textbf{x}_{e}^{T}\textbf{Px}_{e} -\textbf{x}_{e}^{T}\textbf{P}\boldsymbol\Sigma\textbf{Px}_{e}\\ \nonumber
   &=&  \textbf{x}_{e}^{T}\textbf{Px}_{e} - \textbf{x}_{e}^{T}\textbf{P}\boldsymbol\mu+\sigma^{-2}\tilde{\textbf{y}}_{k}^{T}\tilde{\textbf{A}}\boldsymbol\Sigma\textbf{Px}_{e},
\end{eqnarray}
which then gives the log likelihood function in (\ref{eq:log}).

\subsection{Derivation of Update Expressions for Modified RVM}
Firstly, differentiating with respect to $p_{i}$ gives
\begin{equation}\label{eq:p1}
  -\frac{1}{2}\Big[\Sigma_{nn}-\frac{1}{p_{n}}+\mu^{2}_{n}+x_{e,n}^{2}-x_{e,n}\mu_{n} \Big]
\end{equation}
and equating to zero gives
\begin{eqnarray}\label{eq:p2}
  \Sigma_{nn}-\frac{1}{p_{n}}+\mu^{2}_{n}+x_{e,n}^{2}-x_{e,n}\mu_{n} &=& 0 \\ \nonumber
  1 - p_{n}\Sigma_{nn}-p_{n}\mu_{n}^{2}-p_{n}x_{e,n}^{2}+p_{n}x_{e,n}\mu_{n} &=& 0 \\ \nonumber
  \gamma_{n}-p_{n}[\mu_{n}^{2}+x_{e,n}^{2}-x_{e,n}\mu_{n}]&=&0 \\ \nonumber
\end{eqnarray}
which leads to (\ref{eq:pnew}).

Now collect the terms with $\sigma$ in to give
\begin{equation}\label{eq:sig1}
  -\frac{1}{2}\Big[ 2M\log\sigma^{2} - \log|\boldsymbol\Sigma| + \sigma^{-2}||\tilde{\textbf{y}}_{k}-\tilde{\textbf{A}}\boldsymbol\mu||_{2}^{2} \Big]
\end{equation}
and then define $\tau=\sigma^{-2}$ giving
\begin{eqnarray}\label{eq:sig2}
  && -\frac{1}{2}\Big[ 2M\log\tau^{-1} - \log|\boldsymbol\Sigma| + \tau||\tilde{\textbf{y}}_{k}-\tilde{\textbf{A}}\boldsymbol\mu||_{2}^{2} \Big] \\ \nonumber
  &=& -\frac{1}{2}\Big[ -2M\log\tau - \log|\boldsymbol\Sigma| + \tau||\tilde{\textbf{y}}_{k}-\tilde{\textbf{A}}\boldsymbol\mu||_{2}^{2} \Big]. \\ \nonumber
\end{eqnarray}
Now differentiate (\ref{eq:sig2}) with respect to $\tau$ and equate to zero to give
\begin{equation}\label{eq:sig3}
  -\frac{2M}{\tau}+\text{tr}(\Sigma\tilde{\textbf{A}}^{T}\tilde{\textbf{A}})+||\tilde{\textbf{y}}_{k}-\tilde{\textbf{A}}\boldsymbol\mu||_{2}^{2} = 0,
\end{equation}
where $\text{tr}(\cdot)$ indicates the trace.  As $\text{tr}(\Sigma\tilde{\textbf{A}}^{T}\tilde{\textbf{A}})$ can be written as $\tau^{-1}\sum\limits_{n}\gamma_{n}$ we now have
\begin{equation}\label{eq:sig4}
  \tau^{-1}(2M-\sum\limits_{n}\gamma_{n})=||\tilde{\textbf{y}}_{k}-\tilde{\textbf{A}}\boldsymbol\mu||_{2}^{2},
\end{equation}
which in turn gives (\ref{eq:sigmanew}).
\section*{Acknowledgments}

\noindent We appreciate the support of the UK Engineering and Physical Sciences Research Council (EPSRC) via the project Bayesian Tracking and Reasoning over Time (BTaRoT) grant EP/K021516/1.
\balance
\bibliographystyle{IEEEtran}
\bibliography{mybib}

\end{document}